\documentclass{article}

\usepackage{PRIMEarxiv}

\usepackage[utf8]{inputenc} % allow utf-8 input
\usepackage[T1]{fontenc}    % use 8-bit T1 fonts
\usepackage{hyperref}       % hyperlinks
\usepackage{url}            % simple URL typesetting
\usepackage{booktabs}       % professional-quality tables
\usepackage{amsfonts}       % blackboard math symbols
\usepackage{nicefrac}       % compact symbols for 1/2, etc.
\usepackage{microtype}      % microtypography
\usepackage{lipsum}
\usepackage{fancyhdr}       % header
\usepackage{graphicx}       % graphics
\graphicspath{{media/}}     % organize your images and other figures under media/ folder

\usepackage{algorithm}
\usepackage{algorithmic}
\usepackage{subfig}

\usepackage{amsmath}
\DeclareMathOperator*{\argmax}{arg\,max}

\title{Learning to Explore by Reinforcement over High-Level Options
}

\author{
  Juncheng Liu, Brendan McCane, Steven Mills \\
  Department of Computer Science\\
  University of Otago\\
  Dunedin, New Zealand \\
  \texttt{\{juncheng.liu, brendan.mccane, steven.mills\}@otago.ac.nz} \\
}

\begin{document}
\maketitle

\begin{abstract}
Autonomous 3D environment exploration is a fundamental task for various applications such as navigation. The goal of exploration is to investigate a new environment and build its occupancy map efficiently. In this paper, we propose a new method which grants an agent two intertwined options of behaviors: ``look-around'' and ``frontier navigation''. This is implemented by an option-critic architecture and trained by reinforcement learning algorithms. In each timestep, an agent produces an option and a corresponding action according to the policy. We also take advantage of macro-actions by incorporating classic path-planning techniques to increase training efficiency. We demonstrate the effectiveness of the proposed method on two publicly available 3D environment datasets and the results show our method achieves higher coverage than competing techniques with better efficiency.
\end{abstract}

\section{Introduction}
When a robot is placed in a new environment, it is very important that the surroundings are mapped as quickly as possible in an unsupervised manner so that subsequent tasks can be more easily completed.  Specifically, an optimal policy should be able to give a sequence of actions that maximizes the coverage of an environment given a limited time or energy budget. This process is called autonomous exploration and it has been studied for many years.

Most existing work tackles this problem either by active SLAM~(simultaneous localization and mapping) or reinforcement learning. Traditional methods consider the problem as a partially observable Markov decision process~(POMDP)~\cite{white1991survey} while reinforcement learning algorithms, which have attracted more attention in recent years, employ various kinds of rewards such as curiosity, coverage or novelty~\cite{ramakrishnan2021exploration} to encourage the exploration of unknown places.

Compared to traditional visual-SLAM methods, learning-based methods are able to leverage the structural regularities of environments with semantic features. However, one main drawback of reinforcement learning-based exploration is the inefficient use of training data and extremely long training times. This is partly because the action set of an agent is too low-level to train by a limited number of training episodes in an end-to-end manner, though theoretically feasible. One solution is imitation learning  \cite{chen2019learning}, but expert demonstration is usually hard to obtain and hard to interpret. Recently, ~\cite{chaplot2020learning} proposed to use high-level macro-actions instead by incorporating classic path-planning techniques. By doing so, the agent is only trained to select optimal ``goal points'' rather than atomic actions, which significantly improves training efficiency.

In this paper, we propose an autonomous exploration algorithm which integrates two exploration options implemented using the option-critic architecture: navigating to a selected frontier point; and investigating the local environment. An overview of our method is shown in Fig~citetref{pipeline}, and we make the following contributions: 
\begin{itemize}
    \item We show, for the first time, exploration in large spaces using RL across high-level task options that are interleaved in real-time; 
    \item We extend the state-of-the art for exploratory navigation on the Gibson and Matterport3D datasets; and 
    \item We show that our method produces not only more efficient, but qualitatively more compact exploration trajectories.
\end{itemize}

\section{Related work}

\begin{figure*} % picture
    \centering
    \includegraphics[width=1\linewidth]{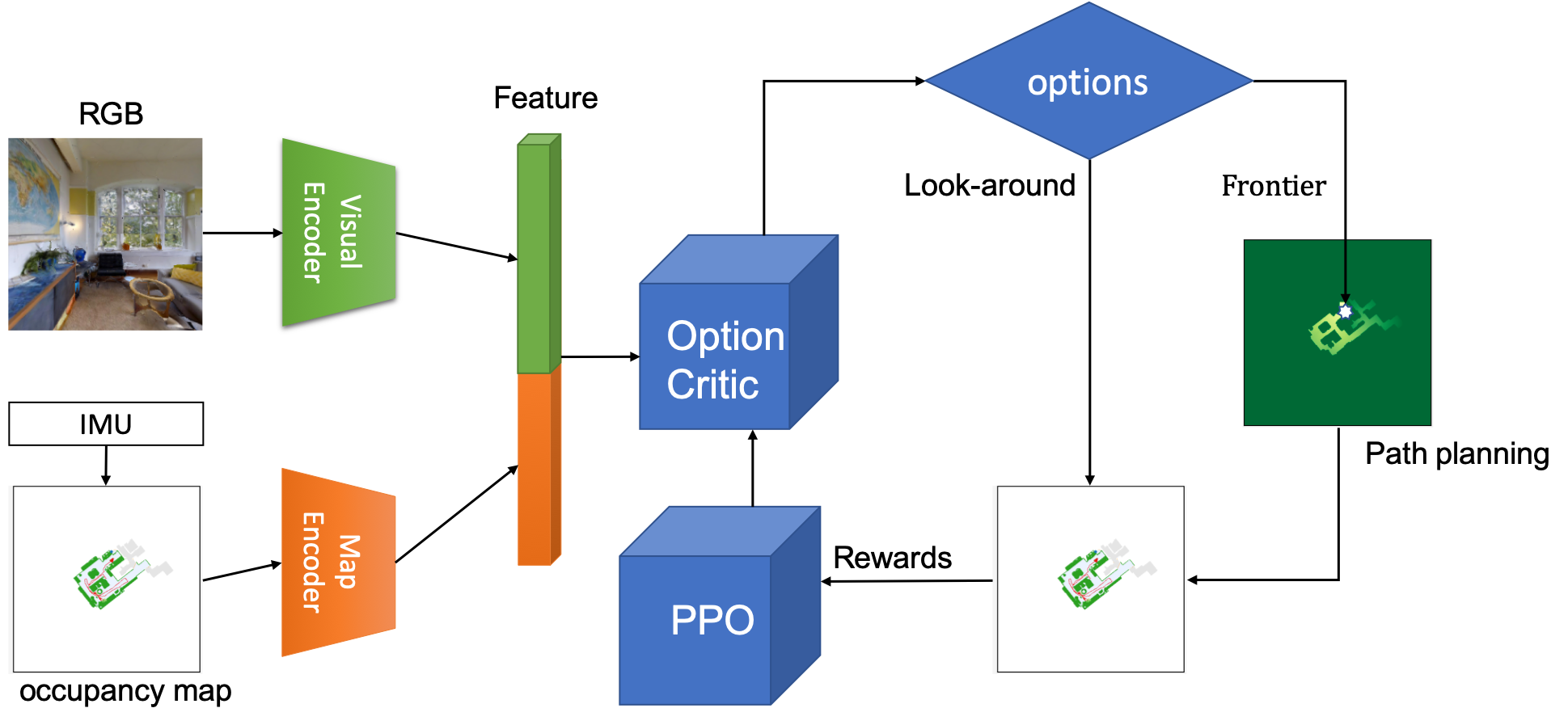}
    \caption{Algorithm overview. Our method takes as inputs the RGB-D frames and the maintained maps. The feature vectors of them are concatenated and fed into the policy network. The policy chooses an option from ``Frontier'' and ``look-around'' according to the estimated values. An action is also produced by the chosen option. To navigate to a frontier, path-planning techniques are employed.}
    \label{pipeline}
\end{figure*}

%\paragraph{Option-critic architecture.} 

~\cite{yamauchi1997frontier} proposed a frontier-based exploration method, that always navigates to the closest frontier point, then always performs a rotation inspection. Subsequently, new frontiers are produced and the process repeats and eventually the whole scene is explored. There is, however, no learning involved.

Recently there have been multiple attempts to solve the autonomous exploration problem by learning. The motivation is that an agent is able to act more efficiently by leveraging the structural regularity and semantics of the scene. Another benefit is by using learning techniques, more freedom of sensory modalities is achieved. For instance, depth information can be inferred by neural networks if only an RGB camera is available. The first end-to-end method to output atomic actions from sensory observation is ~\cite{chen2019learning}. Their policy network takes as inputs an RGB-D frame and the current occupancy map and outputs an action which the agent executes. The reward is designed as the incremental coverage of the occupancy map. This simple method requires imitation learning to initialize the weights of the network. In addition to the spatial memory, the method also integrates a temporal memory by employing GRUs~(gated recurrent units). However, it is still not clear that the method without imitation learning outperforms the classical frontier method \cite{yamauchi1997frontier} . 

While coverage is a common metric, there are also alternative ways of defining rewards: ~\cite{pathak2017curiosity} encourages the agent to visit states that are not predicted confidently, which is also called a curiosity-driven strategy. ~\cite{tang2017exploration} uses the inverse of the square-root of visitation number as the reward, encouraging the agent to move to under-explored areas. We point readers to ~\cite{ramakrishnan2021exploration} for a detailed taxonomy and explanation of existing exploration paradigms.

As a major improvement, ~\cite{chaplot2020learning} leverages a hierarchical structure and classic path-planning techniques to train a policy more efficiently. Instead of using atomic actions, its policy outputs locations where the agent is supposed to navigate to. Compared to atomic actions, higher level actions need fewer training episodes and do not require imitation learning. 
Instead of an actual occupancy map, ~\cite{ramakrishnan2020occupancy} uses an anticipated occupancy map. With training, the anticipated map becomes more and more accurate. This allows the policy access to a complete (albeit approximate) map at a relatively early stage.

All the aforementioned learning methods have only one option which involves navigating to a certain point. Our method combines the frontier method and the line of work which takes advantage of a temporally abstracted policy. However, we employ the option-critic architecture which grants our policy two different options, that is, two options of behaviors, giving the agent more flexibility and efficiency. This surprisingly simple addition of a ``look-around" option, improves the performance of the agent significantly.

\section{Methods}

The goal of autonomous exploration is to establish a 2D occupancy map of an environment as quickly as possible. Given a set of actions $\mathcal{A}$ that an agent can perform, e.g. rotate, move forward, the algorithm should be able to find an optimal trajectory of actions $\tau=\{a_1,a_2,...,a_T|a_i \in \mathcal{A}\}$ such that for a given time step $T$, highest coverage rate of the environment is reached. We solve this problem by estimating a policy, $\pi$, trained by Reinforcement Learning~(RL). 

We consider a mobile agent equipped with an RGB-D camera. For simplicity, we also assume the agent has an IMU module which records the current location relative to the initial position. For a robot without such functionality, some off-the-shelf classic SLAM algorithms~\cite{klein2007parallel,mur2015orb} or neural network-based estimation~\cite{chaplot2020learning} can be applied instead. At each time step $t$, the policy $\pi$ takes as input the current state, $s_t$; an RGB frame; and the current occupancy maps, the format of which will be discussed in detail in the following. The output of the policy is an action $a_t\in \mathcal{A}$ that maximizes the coverage. 

It has recently been found that using higher-level macro-actions significantly improves the training efficiency without involving imitation learning which usually requires a lot of expert demonstrations~\cite{chaplot2020learning}. Similar to this approach, we adopt classic path-planning techniques~\cite{sethian1996fast} for navigation, which saves the agent from learning straightforward navigation tasks. Specifically, instead of outputting an atomic action, our policy $\pi$ estimates a goal point to which the agent will navigate. Different to ANS~\cite{chaplot2020learning} that uses arbitrary points, we incorporate the concept of ``frontiers'' which have been shown as a very effective strategy for exploration~\cite{yamauchi1997frontier}. 

Furthermore, instead of simply performing navigation, we add a complementary option ``look-around investigation''. Our insight is a ``look-around'' operation is sometimes more efficient than wandering around. Imagine when you are in a new environment, the first step you would take is very likely to look around rather than moving to a new location. 

\subsection{Map formats}
It has been demonstrated that complex map-based architectures significantly improve the performance of exploration~\cite{ramakrishnan2020occupancy,henriques2018mapnet}. Furthermore, an explicit map facilitates classic path-planning techniques. We also take advantage of such maps to help memorize the occupancy of environments and trajectory of an agent. Specifically, we use 5 maps to record this information. All of these maps have a dimension of $512\times 512$ and are concatenated as a 5-channel input consisting of: occupancy map, explored map, trajectory map, current location map, frontier map.  
The occupancy map records obstacles. Since our agent has a depth camera, we are able to observe a pointcloud and align it with agent's current location in realtime. Therefore we can build an occupancy map on-the-fly. It is a $512\times 512$ binary map indicating whether a certain location is free or an obstacle. The explored map indicates which areas have been observed, and which are unexplored, while the current location and the past trajectory of the agent are stored in the current location map and trajectory map separately. 

We also maintain a frontier map in realtime. The frontiers are defined as the boundaries between the explored and unexplored areas except for the obstacles (e.g. walls). The computation of this map is very fast since it only involves logical bit-wise operation of the explored and occupancy maps.

For the feature extractor, we use seven 2D convolutional layers with kernel size of 3 and stride of 2 followed by two fully connected layers activated by ReLU. In addition to the maps, the RGB frames are also processed by ResNet18~\cite{he2016deep} and concatenated with the map feature.

\subsection{Training policy}

As previously mentioned, our policy has two options:
\begin{enumerate}
    \item Navigation to a selected frontier, and 
    \item look around at the current location. 
\end{enumerate}
We denote these as $\omega_1$ and $\omega_2$ respectively. At each time step $t$ the policy $\pi$ outputs an option $\omega_t$ as well as the action $a_t$ based on its individual policy $\pi_{\omega_t}$. 

There are three main components in our policy networks parameterized by $\theta,\eta$: the option-value function $V(s,\omega)$, intra-option policies $\pi_{\omega,\theta}$ and termination functions $\beta_{\omega,\eta}$. The option-value function $V(s_t,\omega_t)$ predicts the discounted returns of choosing option $\omega_t \in \{\omega_1,\omega_2\}$ given the current state $s_t$. The intra-option policies $\pi_{\omega,\theta}$ estimate the action distribution in the context of a state and an option. The termination functions $\beta_{\omega,\eta} \in (0,1)$ give probabilities of terminating the current option. Moreover, there is a policy over options $\pi_\Omega$ which selects an option each time the current option is terminated. However we simply use a greedy policy for $\pi_\Omega$, which always picks the option with the largest estimated value: 
\begin{equation}
\pi_\Omega(s_t) = \argmax_{\omega \in \{\omega_1,\omega_2\}} V(s_t,\omega).
\end{equation}

In the following we will present the gradients for intra-option policies $\pi_{\omega,\theta}$ and termination functions $\beta_{\omega,\eta}$ with respect to parameters $\theta$ and $\eta$. Given an option $\omega$ and state $s$, the gradient of an action is similar to that of Advantage-Actor Critic~(A2C)~\cite{mnih2016asynchronous}:
\begin{equation}
%\frac{\partial Q(s,\omega)}{\partial \theta} =
\sum_{a \in \mathcal{A}_\omega} \frac{\partial \pi_{\omega,\theta}(a|s)}{\partial \theta} A(s,\omega, a),
\end{equation}
where $A(s,\omega, a)$ is the ``advantage'' of an action over the averaged rewards:
\begin{equation}
    A(s,\omega, a) = r + \gamma V(s^\prime,\omega) - V(s,\omega),
\end{equation}
$s^\prime$ is the next observation and $\gamma$ is the discount factor. We use $\gamma=0.99$ across all our experiments.

Different to \cite{bacon2017option}, we don't assume the same action space for both options since a navigation target is a location $(x,y)$ while look-around only needs an angle $\alpha$. Therefore we use $\mathcal{A}_\omega$ to denote the action space of option $\omega$.

\begin{algorithm}[tb]
\caption{Exploration with options}
\label{alg:algorithm}
\begin{algorithmic}[1] %[1] enables line numbers
\STATE Choose the initial option $\omega_0$ according to $\pi_\Omega(s_0)$.
\STATE $\omega\leftarrow \omega_0$, $s \leftarrow s_0$
\FOR{$i=1,...,MaxSteps$}
\STATE \textbf{Evaluation stage:}
\STATE Generate a macro-action $a\sim \pi_{\omega,\theta}(s)$
\IF{$\omega$ is ``Frontier-navigation''}
\STATE Navigate to $a$ in Environment using Path Planning and obtain $s^\prime$, $r$
\ELSE
\STATE Rotate by an angle $a$ in Environment and obtain $s^\prime$, $r$
\ENDIF
\STATE $s \leftarrow s^\prime$
\STATE Calculate $\beta_1$, $\beta_2$ and $V(s,\omega_1)$, $V(s,\omega_2)$.
\IF {Bernoulli($\beta_{\omega,\eta}$)}
\STATE Update $\omega \leftarrow \pi_\Omega(s)$
\ENDIF
\STATE \textbf{Training stage:}
\STATE $\theta \leftarrow
\frac{\partial \log \pi_{\omega,\theta}(a|s)}{\partial \theta} A(s,\omega, a)$\\
\STATE $\eta \leftarrow
\frac{\partial \beta_{\omega,\eta}(s)}{\partial \eta} ( V(s,\omega) - \max_{\omega^\prime} V(s,\omega^\prime))$
\STATE Update $\theta$ and $\eta$ based on Eq\ref{eq:tderror}.
\ENDFOR
\end{algorithmic}
\label{algorithm}
\end{algorithm}

The gradient for $\beta_{\omega,\eta}$ with respect to $\eta$ is as follows:

\begin{equation}
\sum_{\omega \in \{\omega_1,\omega_2\}} \frac{\partial \beta_{\omega,\eta}(s)}{\partial \eta} ( V(s,\omega) - \max_{\omega^\prime} V(s,\omega^\prime)),
\end{equation}
where $V(s,\omega) - \max_{\omega^\prime} V(s,\omega^\prime)$ acts in a similar way to advantage, which increases the termination probability of an option when its estimated value is suboptimal.

Finally, we update the value function parameters according to the TD-target:
\begin{equation}
r + \gamma ((1-\beta) V(s',\omega) + \beta \max_{\omega^\prime} V(s,\omega^\prime)).
\label{eq:tderror}
\end{equation}

\begin{figure} % picture
    \centering
    \includegraphics[width=0.85\linewidth]{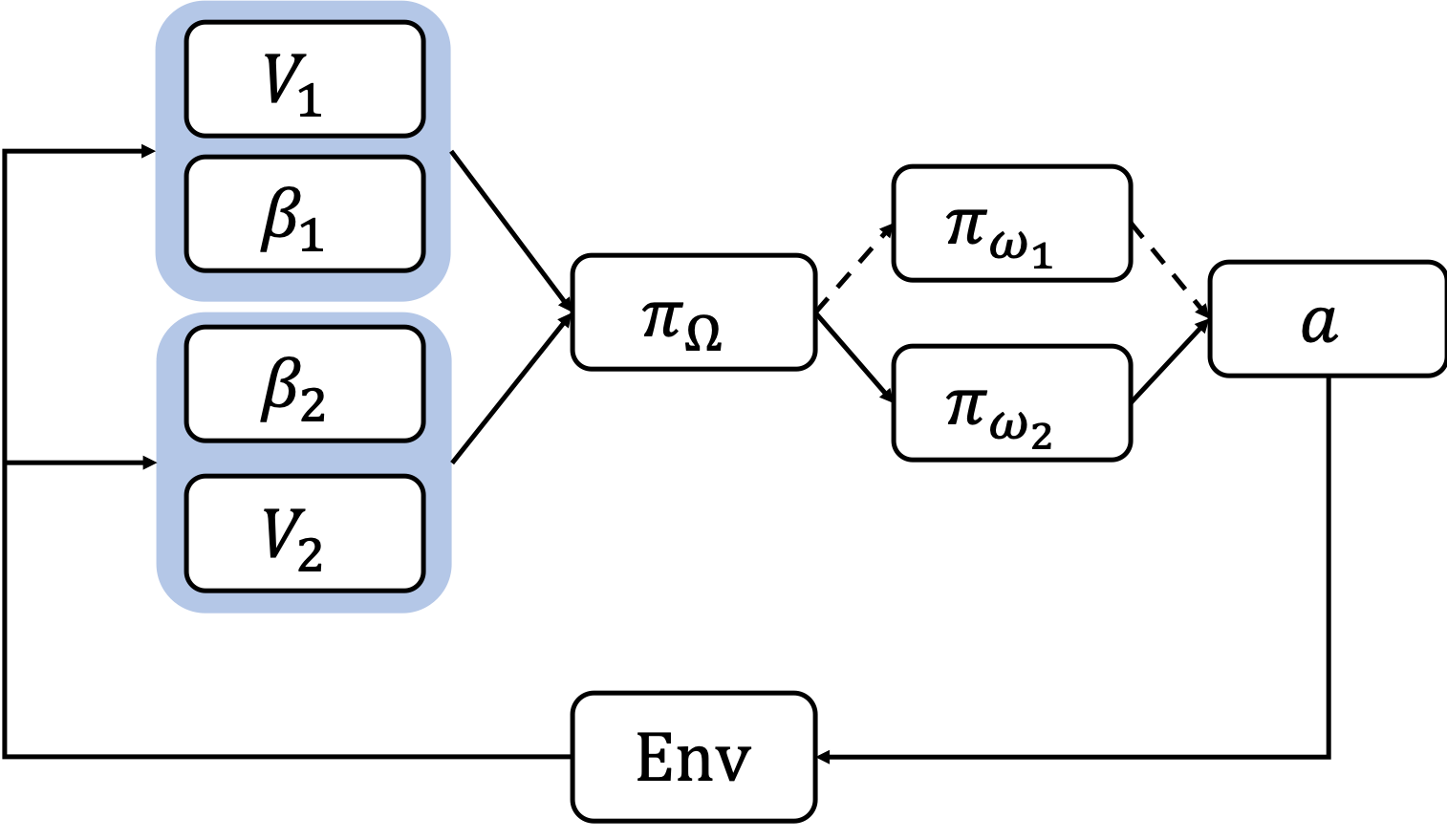}
    \caption{Model diagram. After a macro-action is executed, values $V$ and terminations $\beta$ are computed. The next option is decided accordingly. }
    \label{fig:diagram}
\end{figure}

Our proposed approach is summarized in Algorithm~\ref{alg:algorithm} and its architecture is shown in Fig~\ref{fig:diagram}. The method is divided into training and evaluation stages. During training, we employ a memory buffer which stores 20 different trajectories. The policy is then trained by these accumulated trajectories.

\subsection{Rewards}
Similar to \cite{chen2019learning,chaplot2020learning}, we adopt the increase of coverage during exploration as rewards. This reward indicates how much new knowledge an action gains from the environment. The new explored area can either be free space or obstacles. And is also an evaluation of how efficient an action is. We use the ratio of increment to the total area of a scene instead of the absolute area considering that different scenes might have different scales.

\section{Experiments}
\subsection{Experimental setup}
We evaluate our proposed method in Habitat-Lab, a modular high-level library for end-to-end development in embodied AI~\cite{savva2019habitat} on two publicly available datasets: Gibson~\cite{xia2018gibson} and Matterport3D~\cite{chang2017matterport3d}. Both the datasets consist of 3D reconstructions of real-world indoor environments such as offices and homes. The average area of scenes in Matterport3D is larger than Gibson. For both datasets, there are training/test splits available. We use the \emph{val} split for testing Gibson and the \emph{test} set for Matterport3D. For each training/test scene (an episode), we use a total number of 1000 steps.   

Although we make assumptions about the input modalities, we note that our method is not limited to a robot with such a configuration. The flexibility of input modalities can be achieved by adopting neural network-based approximators. However, since this is not the main focus of the paper, we simply use the ground-truth data of the simulator. For each timestep, the agent observes an RGB-D frame of size $256\times 256$ and its location. The aforementioned maps are maintained and updated in realtime using this information. 

Our algorithm is implemented in PyTorch~\cite{NEURIPS2019_9015}. All experiments were done on a PC with Intel Core i7-6700K CPU (4.00GHz) and an NVIDIA Quadro P6000 GPU (24GB memory).

\subsection{Baselines}

\begin{figure*}[t] % picture
    \centering
    \includegraphics[width=1\linewidth]{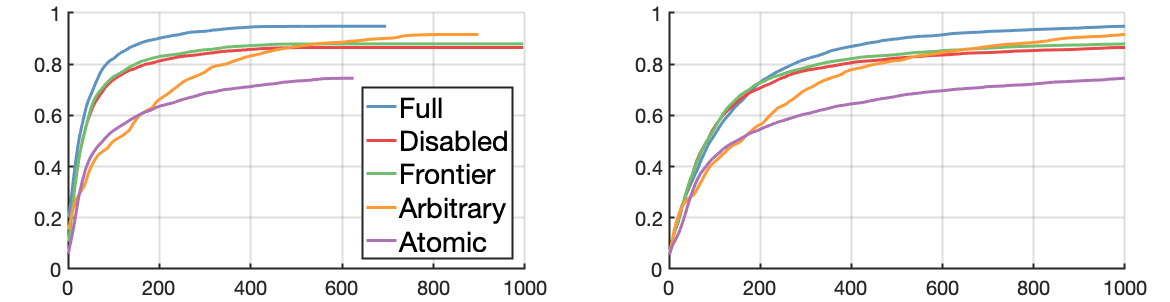}
    \caption{Evaluation on Gibson. Plots show the averaged coverage as the episodes progress. Left: coverage with trajectory-length. Right: coverage with timesteps. }
    \label{fig:gibson}
\end{figure*}

\begin{figure*}[t] % picture
    \centering
    \includegraphics[width=1\linewidth]{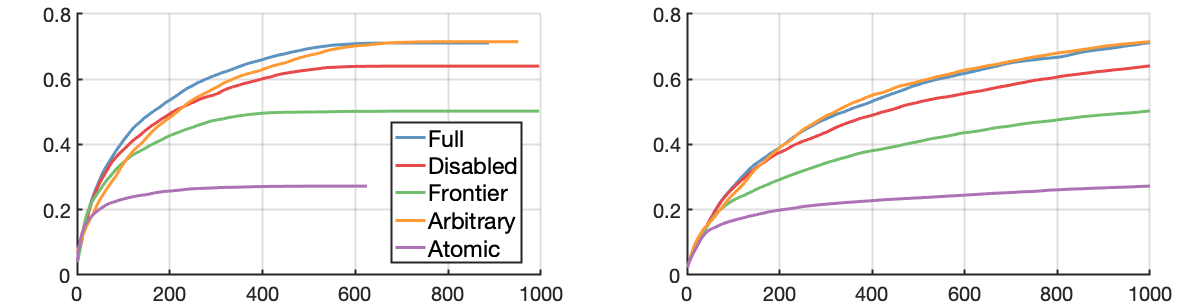}
    \caption{Evaluation on Matterport3D. Plots show the averaged coverage as the episodes progress. Left: coverage with trajectory-length. Right: coverage with timesteps.}
    \label{fig:mp3d}
\end{figure*}

We use 4 alternative baselines for evaluation and comparisons. All of the baselines as well as our method are trained with the same 750 episodes~(except for the frontier method which does not need training) and tested on the same split. It's worth noting that the test scenes are not seen during training and we do not train the policy during testing. To speed up the training process, we use 8 simulators to collect trajectories in parallel. Since we employ the A2C~\cite{mnih2016asynchronous} algorithm, every simulator is synchronized when executing actions. We use 50 steps for all macro-actions: frontier-navigation and look-around. The policy is updated by the collected trajectories after every 20 macro-actions are executed using proximal policy optimization~\cite{schulman2017proximal}. 

\begin{table}[t]
\centering
%\resizebox{.95\columnwidth}{!}{
\begin{tabular}{l|l|l|l|l}
    Methods & Gib500 & Gib1k & Mat500 & Mat1k \\
    \hline
    Full & \textbf{0.90} & \textbf{0.95} & 0.58 & \textbf{0.71}\\
    Disabled & 0.82 & 0.87 & 0.53 & 0.64 \\
    Frontier & 0.84 & 0.88 & 0.41 & 0.50 \\
    Arbitrary & 0.82 & 0.92 & \textbf{0.59} & \textbf{0.71}\\
    Atomic & 0.68 & 0.75 & 0.24 & 0.27\\
       
\end{tabular}
\caption{Coverage after 500 and 1000 steps on Gibson and Matterport3D datasets.}
\label{tab:table}
\end{table}

\paragraph{Frontier method.} This is the classic method for exploration without any learning being involved~\cite{yamauchi1997frontier}. The main idea is to chase after the closest frontiers until the whole environment is explored. The simple policy performs well and is guaranteed to converge given enough time.

\paragraph{Atomic action method.} The simplest attempt to engage RL to the problem is just maximizing the coverage given a set of atomic actions (\emph{turn left, turn right, move forward}) such as \cite{chen2019learning}. To make the comparison straightforward, we do not employ imitation learning as done in  \cite{chen2019learning}. Furthermore, we assume the agent is provided with the ground-truth locations. 

\paragraph{Arbitrary point method.} Instead of navigating to frontier points, \cite{chaplot2020learning} proposed to navigate to arbitrary points on a map. Similarly, we also simplify its original implementation for the straightforwardness of comparison: we do not include SLAM and local policy components. 

\paragraph{Option disabled.} To validate our method benefits from the look-around option, we perform an experiment with the look-around option disabled. 

The major difference between our proposed method and the alternatives is our method has two options~(navigation and look-around) while the others only have one.

\subsection{Metrics}
We use two metrics for evaluating the performance of all methods: coverage percentage with timesteps and coverage percentage with trajectory length. The difference between them is that trajectory length only considers the number of ``\emph{move forward}'' actions while timesteps take all of the three actions into consideration. The length provides an effective way of measuring the neatness and efficiency of a trajectory. Besides, for a robot equipped with a rotatable camera, rotating its arm usually costs less energy than moving forward.

\subsection{Results and comparisons}

The results for Gibson and Matterport3D datasets are shown in Fig~\ref{fig:gibson} and Fig~\ref{fig:mp3d} respectively. Overall, our method achieves the best performance in both metrics and datasets. After a total number of 1000 time steps, our method achieves a coverage of 95\% across all test scenes of Gibson on average, followed by 92\% for the arbitrary point method, 88\% for the frontier method. The atomic action method achieves the lowest coverage rate of 75\%. The ranking of the  methods for Matterport3D closely agrees with that of Gibson. 

There are 14 test scenes in total which are not seen during training. Each scene has 71 different 2D rotations and initial locations of agent for Gibson. In Matterport3D, there are 18 test scenes in this dataset in total and each one has 51 different rotations. The scenes in this dataset are generally much larger than the ones in Gibson and their layouts are more complicated, which brings more difficulty to the exploration given the same number of time steps. 

Table~\ref{tab:table} collects the coverage rates of the five aforementioned approaches after 500 and 1000 timesteps in the two datasets respectively. It can be seen that for the same method and number of steps, the coverage of Matterport3D is lower than that of Gibson by approximately 20\%-30\%. When considering the coverage with trajectory length (left of Fig~\ref{fig:mp3d}), our method is slightly better than arbitrary point method especially in the initial stage. 

The results suggest that temporally abstracted macro-actions are generally more effective than atomic actions. This is due to the fact that classic path planning is employed to obtain optimal paths, while learning this usually requires much more training. By using high-level behaviors the policy focuses on goals with higher abstraction level. We note that the frontier-based method achieves decent coverage without learning. Especially at the beginning stage, it even outperforms the arbitrary point method. However, with the exploration progressing, the learning-based method is able to explore more areas as it learns the regularities from the past trajectories and observations.

We observe that the trajectories produced by our method appear more organized and cleaner than that of the frontier method. This is the consequence of the option switching and planned frontier goals. For circumstances where a look-around action is more efficient than navigating to a point, the option-critic architecture enables switching between multiple options, leading to a shorter trajectory length. 

We show the exploration trajectories in Fig~\ref{fig:trajectories}. This visually validates the neatness of the trajectories of our method. Our method is advantageous as it uses fewer \emph{move forward} actions which are replaced by rotation wherever possible.

\begin{figure} % picture
\begin{center}
	\subfloat[Matterport3D small.]{
		\label{fig:mp3d-small}
		\centering
		\includegraphics[width=0.5\linewidth]{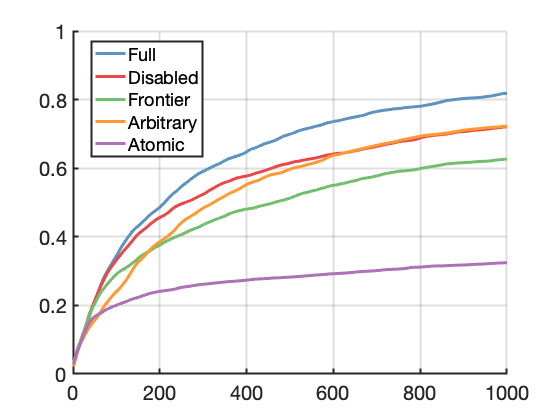}
	}
	\subfloat[Domain generalization performance. Coverage with trajectory-length. We evaluate the performance on Matterport3D with the model trained on Gibson.]{
		\centering
		\includegraphics[width=0.5\linewidth]{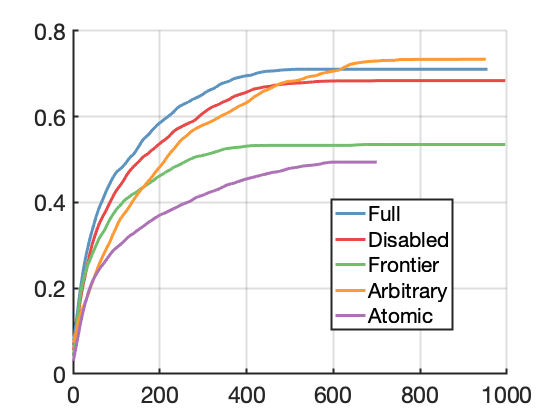}
        %\caption{Coverage with trajectory-length. We evaluate the performance on Matterport3D with the model trained on Gibson.}
        \label{fig:transfer}
	}
\end{center}
\end{figure}

Since the areas of scenes in this dataset are more varied, the performance for scenes of different scales exhibits relatively large variety. We observe that our method is most advantageous in small to medium scenes. In Fig~\ref{fig:mp3d-small} we show the coverage with timesteps with the smallest 50\% of scenes in Matterport3D. Compared to the right subfigure of Fig~\ref{fig:mp3d}, on smaller scenes, our method outperforms the others by a larger margin. 

\subsection{Ablations}
\begin{figure} % picture
    \centering
    \includegraphics[width=0.9\linewidth]{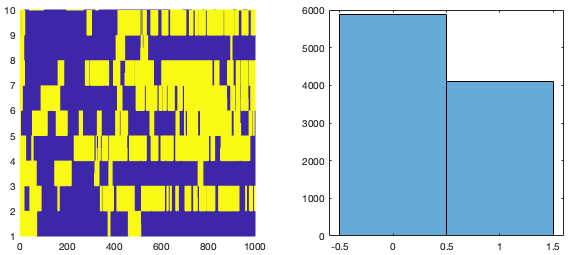}
    \caption{Option selection and histogram. Left: 10 option selections in 10 trajectories. The x-axis shows the timesteps and the y-axis has a row for each trajectory. Navigation is shown as purple and Look-around yellow. Right: frequency of options. Navigation is 0, look-around is 1.}
    \label{fig:hist}
\end{figure}

\begin{figure*}[t] % picture
    \centering
    \includegraphics[width=1\linewidth]{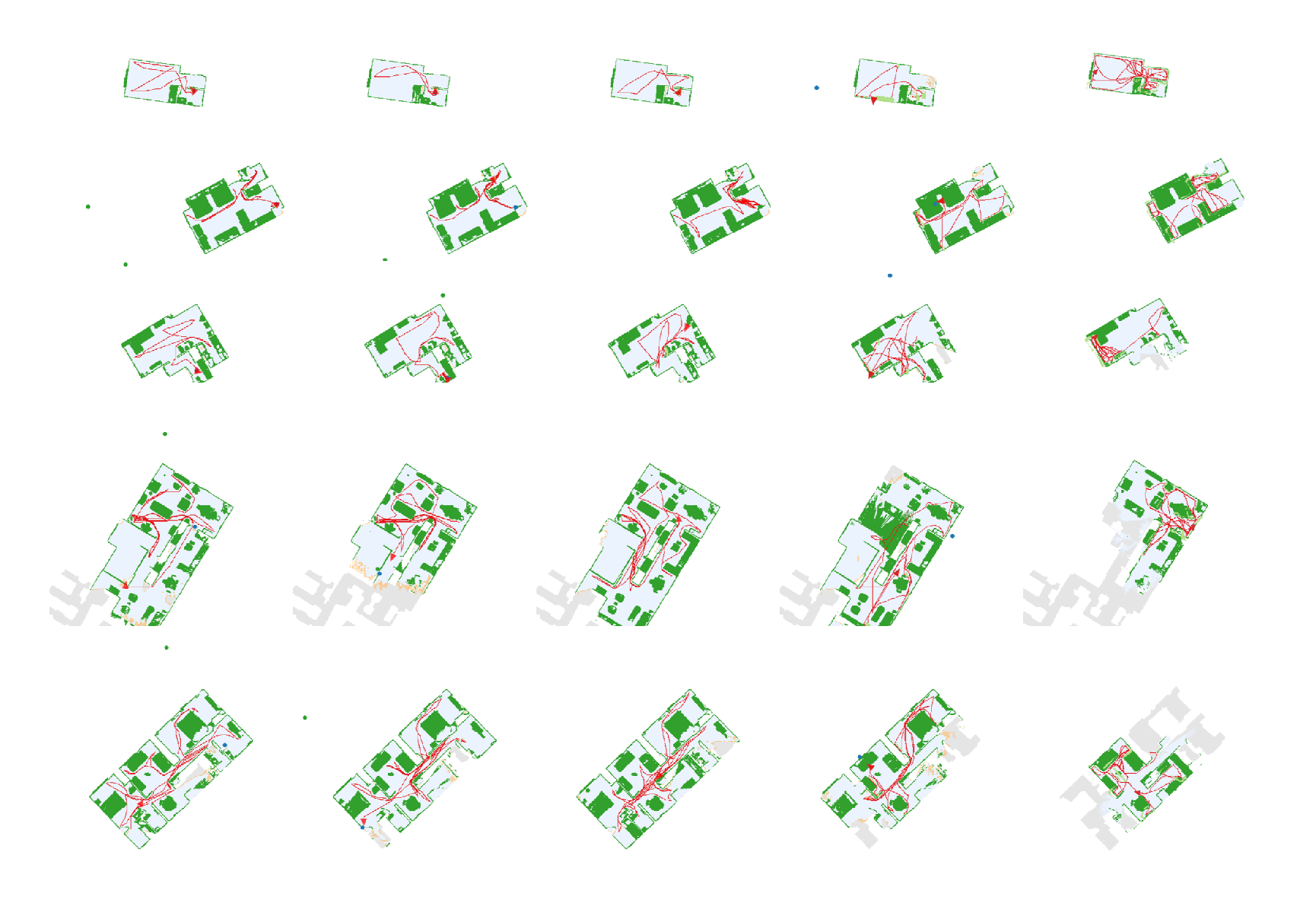}
    \caption{Trajectories of different methods after 1000 timesteps. From left to right columns: our approach, look-around option disabled, frontier method, arbitrary point method and atomic action method. Each row represents a selected scene. Green areas represent obstacles while light blue represents free space. Red lines indicated the trajectories of agent. Orange dots are the frontiers. Unexplored areas are in gray.}
    \label{fig:trajectories}
\end{figure*}

To validate our method benefits from both options, we did an ablation study in which only the navigation option is enabled and look-around is disabled. We can see from Fig~\ref{fig:gibson} and \ref{fig:mp3d} that when only navigation is enabled, the performance drops both timestep-wisely and trajectory length-wise. This is because our policy plans with two options in mind. When look-around option is disabled, extra navigation steps are used instead leading to inefficiency. It can also be observed from Fig~\ref{fig:trajectories} that the trajectories of the method with look-around disabled are generally more messy and the coverage is less after the same number of time steps. We show the selections and distribution of options from 10 distinct trajectories in Fig~\ref{fig:hist}. Generally speaking, the navigation option is selected more frequently than look-around. A common pattern of behavior that we observe across different scenes is that the agent starts exploring by first looking around. When most of the scene is explored, the agent tends to choose look-around more often as shown in left of Fig~\ref{fig:hist}.

We also evaluate the generalization ability of the above methods. In this experiment, every method is tested on Matterport3D dataset with the model trained on Gibson. Since the frontier method does not involve learning, it remains the same in both experiments. The purpose is to test if a method can generalize well from one domain to another. The result can be seen in Fig~\ref{fig:transfer} which shows the coverage-with-trajectory-length. It closely agrees with the results shown in left of Fig~\ref{fig:mp3d}. This suggests all the methods generalize well. The arbitrary point method seems to perform better as the exploration finishes.  

\section{Conclusion}
In this paper, we proposed a method that integrates two options for autonomous exploration. One option is navigating to a selected frontier using classic path-planning techniques. The other complimentary option is looking around the surroundings by a certain angle. We validated its effectiveness on two publicly available datasets and the results out-perform the four baselines we tested. We also validated the method benefit from both options by an ablation study.

To conclude, we have the following findings in our research: it is more effective and efficient to have multiple options for exploration. Note that our method is able to integrate any number of options even though we only used two in our implementation. Classic frontier-based method is more advantageous in the early stage. Employing path-planning techniques requires fewer episodes, making training more efficient. Temporally abstracted actions are generally more effective than atomic actions. All of these are consistent across all our experiments.

Future work includes integrating more options and explore how these options can cooperate and what is the minimum set of options that achieves the best performance. Another further direction is a policy outputting a higher hierarchical abstracted action such as a whole planned trajectory. We expect this to take longer-term planning into consideration which makes the trajectory even shorter hence more action-efficient.

%Bibliography
\bibliographystyle{unsrt}  
\bibliography{references}

\end{document}